\documentclass[a4paper]{article}

\usepackage{INTERSPEECH2022}
\usepackage{multirow}
\usepackage{multicol}
\usepackage{graphicx, enumitem}
\usepackage{cite} 
\usepackage{xcolor}
\usepackage{tablefootnote}
 
\newcounter{magicrownumbers}
\newcommand\rownumber{\stepcounter{magicrownumbers}\arabic{magicrownumbers}}
\makeatletter

\title{Exploring linguistic feature and model combination for speech recognition based automatic AD detection}
\name{Yi Wang, Tianzi Wang, Zi Ye, Lingwei Meng, Shoukang Hu, Xixin Wu, Xunying Liu, Helen Meng}
\address{
  The Chinese University of Hong Kong, Hong Kong SAR, China}
\email{\footnotesize{\{ywang,twang,zye,lmeng,skhu,wuxx,xyliu,hmmeng\}@se.cuhk.edu.hk}}

\begin{document}
\bstctlcite{IEEEexample:BSTcontrol}

\maketitle
\vspace{-0.3cm}
\begin{abstract}
\vspace{-0.1cm}
Early diagnosis of Alzheimer’s disease (AD) is crucial in facilitating preventive care and delay progression. Speech based automatic AD screening systems provide a non-intrusive and more scalable alternative to other clinical screening techniques. Scarcity of such specialist data leads to uncertainty in both model selection and feature learning when developing such systems. To this end, this paper investigates the use of feature and model combination approaches to improve the robustness of domain fine-tuning of BERT and Roberta pre-trained text encoders on limited data, before the resulting embedding features being fed into an ensemble of backend classifiers to produce the final AD detection decision via majority voting. Experiments conducted on the ADReSS20 Challenge dataset suggest consistent performance improvements were obtained using model and feature combination in system development. State-of-the-art AD detection accuracies of 91.67\% and 93.75\% were obtained using manual and ASR speech transcripts respectively on the ADReSS20 test set consisting of 48 elderly speakers.
\end{abstract}

\vspace{-0.2cm}
\section{Introduction}
\vspace{-0.1cm}
Alzheimer’s disease (AD), the most common form of dementia often found in aged people, is characterized by progressive degradation of the memory, cognition, and motor skills, and consequently decline in the speech and language skills of patients \cite{sachdev2014classifying, vipperla2010ageing}. Currently, there is no effective cure for AD \cite{Yadav}, but a timely intervention approach can delay its progression and reduce the negative physical and mental impact on patients \cite{de2019brief}. To this end, speech-based automatic AD diagnosis has advantages of being non-intrusive, more scalable and less costly, when compared with conventional approaches based on clinical tests. In recent years there has been a burgeoning interest in developing such systems, notably represented by the recent ADReSS challenges \cite{luz2020alzheimer, luz2021detecting}.

Current speech-based AD detection systems consist of front-end feature extraction and back-end classification components. Commonly used forms of acoustic features extracted from the audio include, but not limited to, spectral features, prosodic features, vocal quality, and pre-trained deep neural network (DNN) embedding features \cite{balagopalan2020bert, syed2020automated, edwards2020multiscale, martinc2020tackling, koo2020exploiting, syed2021tackling, rohanian2021alzheimer, pappagari2021automatic, pan2021using}. Handcrafted linguistic features \cite{santos2017enriching, ammar2018speech, chien2018assessment, balagopalan2020bert, martinc2020tackling} capturing lexical or syntactical cues, or those pre-trained text neural embedding \cite{balagopalan2020bert, syed2020automated, edwards2020multiscale, baidu_paper, pompili2020inesc, qiao2021alzheimer} can be extracted from the transcripts of AD assessment speech recordings. To fully automate the AD detection process, ASR systems \cite{ammar2018speech, chien2018assessment, li2021comparative, gosztolya2019identifying, qiao2021alzheimer, pan2021using, syed2021tackling, zhu2021wavbert, rohanian2021alzheimer, pappagari2021automatic, chen2019attention} have been increasingly often used to produce speech transcripts, in place of ground truth manual transcripts \cite{santos2017enriching, balagopalan2020bert, syed2020automated, baidu_paper, searle2020comparing, sarawgi2020multimodal, li2021comparative} that are costly to annotate in large quantities. In addition, disfluency measures of speech were found useful for AD detection tasks \cite{chien2018assessment, baidu_paper, pappagari2020using, qiao2021alzheimer, rohanian2021alzheimer}. Diverse forms of back-end classifiers range from DNN variants \cite{chen2019attention,  cummins2020comparison, pompili2020inesc, rohanian2021alzheimer} to SVM, decision trees and further combination among these \cite{de2020artificial}. Recent researches have widely reported state-of-the-art AD detection accuracy performance can be obtained using linguistic features \cite{balagopalan2020bert, syed2020automated, baidu_paper, li2021comparative, ye2021development, syed2021automated, qiao2021alzheimer}. Some of them further used model ensembling methods to exploit linguistic features and enhance integration among inner-modal information \cite{baidu_paper, syed2020automated, syed2021automated, qiao2021alzheimer}.

One major challenge confronting the development of automatic speech-based AD detection systems is the scarcity of the required specialist data due to the difficulty in collecting large quantities of transcribed AD assessment speech recordings and shortage of clinical professionals. The resulting “small data” learning task leads to model uncertainty encountered in multiple stages of system development. These include, but not limited to, the following considered in this paper: a) cross-domain transfer text embedding DNNs, for example, state-of-the-art BERT \cite{devlin2018bert} and Roberta \cite{liu2019roberta} pre-trained models, by parameter fine-tuning on limited in-domain AD detection data, leading to over-confident parameters; b) possible errors in the transcripts produced by ASR systems; and c) the suitable form of back-end classifier design and parameter estimation. 

To this end, feature and ensemble approaches are adopted in this paper as overarching solutions to address the above model uncertainty issues. First, the robustness of domain fine-tuning of BERT and Roberta pre-trained text encoders on limited AD detection speech transcripts is improved using model averaging. Second, the uncertainty over the suitable form of text embedding features is accounted for using both BERT and Roberta text embedding representations for back-end classifier construction. Third, the uncertainty over ASR transcript’s quality is further handled using two-pass decoding based speech recognition system combination, or later fusion of their respective AD detection outputs. Lastly, the uncertainty over the back-end classifier is addressed using decision voting among an ensemble of classifiers.

Experiments conducted on the ADReSS20 Challenge dataset suggest consistent performance improvements were obtained using model and feature combination in system development. State-of-the-art AD detection accuracies of 91.67\% and 93.75\% were obtained using manual and state-of-the-art hybrid CNN-TDNN ASR system produced speech transcripts respectively on the ADReSS20 test set consisting of 48 elderly speakers. 

The main contributions of this paper are summarized below. First, this paper presents a systematic application of feature and model combination approaches to address the inherent data sparsity and model uncertainty issues in speech recognition based AD detection systems. In contrast, prior research primarily investigated model combination over back-end classifiers \cite{baidu_paper, syed2020automated, syed2021automated, qiao2021alzheimer, sarawgi2020multimodal, martinc2020tackling}, while its application to cross-domain fine-tuning of pre-trained text embedding and ASR system outputs fusion remains limited. Second, this paper presents a state-of-the-art AD detection accuracy of 91.67\% obtained using manual transcripts and best published accuracy of 93.75\% using ASR speech transcripts respectively on the ADReSS20 test set. To the best of our knowledge, the closest published accuracy performance to date on the same task are 91.67\% \cite{syed2021automated} and 87.50\% \cite{li2021comparative} using manual and ASR speech transcripts, respectively. 

The rest of this paper is organized as follows. Section 2.1 introduces the AD detection task and the ADReSS20 \cite{luz2020alzheimer} dataset used in this work. Section 2.2 briefly describes the hybrid CNN-TDNN and end-to-end Conformer ASR systems, BERT and Roberta text embedding models, and back-end classifiers. Section 3 suggests an approach to fine-tune the pre-trained text encoders. Section 4 presents AD detection systems with feature combination and classifier combination methods on manual transcripts, followed by Section 5 showing AD detection with those methods using ASR system outputs. Finally, the conclusions are drawn, and future works are discussed in Section 6.

\vspace{-0.2cm}
\section{Task Description}
\vspace{-0.1cm}
This subsection describes the audio and text data used in this paper and the baseline system structure.
\vspace{-0.3cm}
\subsection{Dataset}
\vspace{-0.1cm}
In this paper, the ADReSS20 Challenge dataset \cite{luz2020alzheimer} is used for the AD detection system training and evaluation. It is selected from the Cookie Theft picture description part of the Pitt Corpus in the DementiaBank database \cite{becker1994natural} with balanced age, gender and AD label distribution among subjects. The data consists of speech recordings,  corresponding manual transcripts annotated by the CHAT coding system \cite{macwhinney2000childes} and binary AD labels. Both the participant and the interviewer parts are provided. It is divided into the training set of 2 hours 9 minutes from 108 subjects and the test set of 1 hour 6 minutes from 48 subjects.
\vspace{-0.3cm}
\subsection{Baseline System}
\vspace{-0.1cm}
A baseline tandem BERT text embedding and SVM classifier based system structure was used \cite{li2021comparative} to produce AD detection results with either ground truth manual transcripts or ASR system generated transcripts. 
\\
\textbf{ASR Systems:} The hybrid CNN-TDNN system is based on a recently proposed lattice-free MMI trained factored TDNN system \cite{gulanticonformer2020} with additional 2-dimensional convolutional layers being employed as the first 6 layers, and is further optimized with data re-segmentation, domain adaptation, and Bayesian LHUC speaker adaptation\cite{xie2019blhuc}. The end-to-end ASR system used is a Conformer specially developed for elderly speech by training with augmented elderly speech data, incorporating manual re-configuration of Conformer hyper-parameters and cross-domain adaptation. Two-pass decoding based system combination ASR outputs involving the CNN-TDNN ASR system to produce initial N-best recognition hypotheses before being rescored using the Conformer system using a 2-way cross-system score interpolation \cite{tianzi2022conformer, mingyu2022twopass}  are also considered. 
\\ 
\textbf{BERT and Roberta text embedding:}
we combine all participant speech segments corresponding to a single picture description for extracting text features. The principal 768-dimensions of the last hidden layer BERT or Roberta model output are taken as the single vector representation for each participant’s AD classification. This is then normalized by removing mean and scaling to unit variance before being passed into the classifiers. The BERT pre-trained model ${\tt bert-base-uncased}$ is based on the open-source model library \footnote{https://huggingface.co/bert-base-uncased}, with 12 transformer blocks, 768 hidden layers, and 12 attention heads per attention layer. The max input length is 512. The standard BERT tokenizer transforms input texts into tokens. This paper also explored the use of Roberta pre-trained text feature embedding model and the accompanying tokenizer\footnote{https://huggingface.co/roberta-base}. 
\\
\textbf{Classifier:} The SVM classifier used SVC API from the Sklearn package, with the regularization parameter $C$ for soft margin set to 1 and linear kernel as the kernel function. The other four classification models used are Linear Discriminant Analysis (LDA), Gaussian Process (GP), Multilayer Perceptron (MLP)\footnote{https://scikit-learn.org/stable/}, and extreme gradient boost (XGB) \footnote{https://xgboost.readthedocs.io/en/latest/python/index.html}. Each classifier is optimized with the best hyper-parameters found via greedy search on cross-validation data using BERT derived text features of manual transcriptions. The best-performed classifier hyper-parameters are adopted and fixed in all the experiments of this paper. We evaluated the system performance by 10-fold cross-validation (CV)\footnote{In ADReSS20 dataset, each participant only has one recording, so CV is conducted on participant-level} on the training set and tested on the test set. 
\begin{figure*}[t]
\vspace{-1.6cm}
  \centering
  \includegraphics[width=\linewidth]{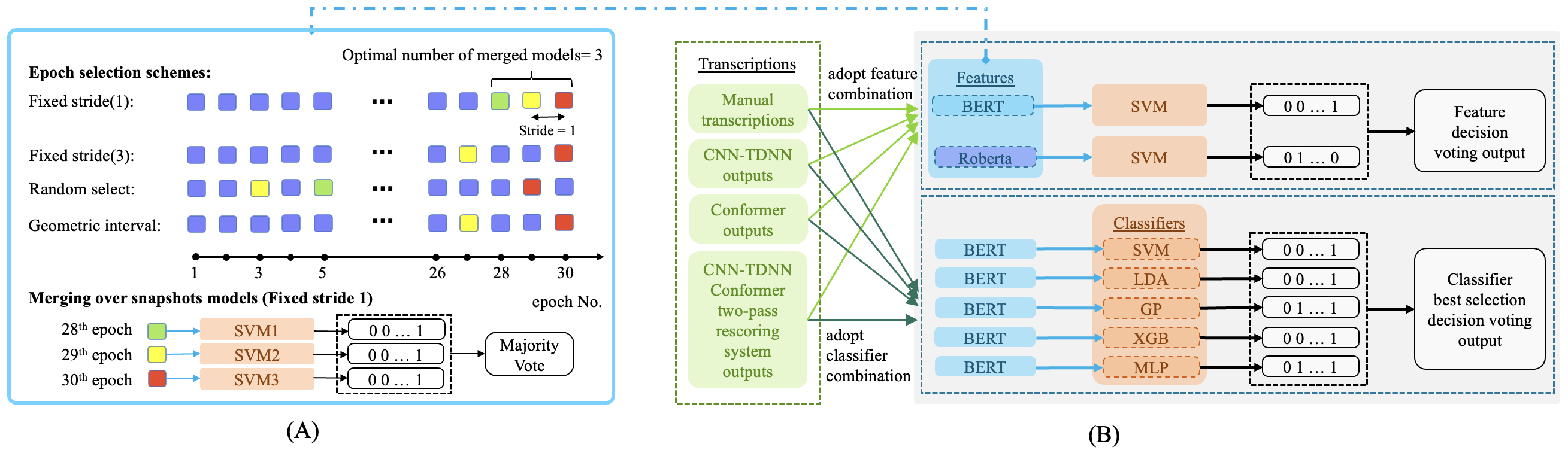}
  \caption{Example of (A) an ensemble of 3 fine-tuned BERT/Roberta pre-trained models on the ADReSS training data obtained at different update epochs (top, left) and being used to produce separate text embedding features for SVM based AD detection outputs majority voting (in green/yellow/red, bottom, left); and (B) AD detection system pipeline using manual, or one of three ASR systems (CNN-TDNN, Conformer, or their combination via two-pass recognition \cite{tianzi2022conformer}) produced transcripts (green) for text embedding features extraction (blue) and back-end classification (orange). Feature combination feeds fine-tuned BERT and Roberta text embedding of (A) into a single back-end SVM classifier for outputs voting (top right).  Classifiers combination feeds fine-tuned BERT or Roberta text features into five classifiers for outputs voting (bottom right). }
  \vspace{-0.2cm}
  \label{fig:AD_detection}
\vspace{-0.3cm}
\end{figure*}
\vspace{-0.2cm}
\section{Fine-tuning Pre-trained Text Encoder}
\vspace{-0.1cm}
To reduce the mismatch between pre-trained BERT/Roberta models built on general texts and elderly speech transcripts, fine-tuning was performed on the 11591 words manual transcripts of the ADReSS training set. We fine-tuned the BERT or Roberta models with the Masked Language Modelling and Next Sentence Prediction tasks used in pre-training. To circumvent the risk of over-fitting, a selected set of three fine-tuned pre-trained models on the ADReSS training data obtained at the final three update epochs were used to produce separate text embedding features for SVM based AD detection outputs majority voting, as is shown in Figure \ref{fig:AD_detection} (A). Ablation studies on alternative update epoch selection schemes, for example, based on random epoch sampling, larger or non-even inter-epoch intervals, suggest the above provides the best AD detection performance in practice. The performance of their corresponding AD detection accuracy on the training set 10-fold CV and the test set are shown in Table \ref{tab:1}. Notably, during the evaluation, we aggregated the scores from ten runs of CV by performing decision voting from models trained in ten runs, as it generated higher stability and reduced the uncertainty in model selection. Accuracy scores are primarily used to evaluate task performance.

Several trends can be found in Table 1. First, irrespective of the precise form of epoch selection being used prior to their combination shown in Figure 1, a trend is that performance improvements were obtained over the non-fine-tuned baseline BERT models, and the directly fine-tuned models without inter-epoch combination (Sys.3 vs. Sys.1-2). Second, epoch selection using a fixed stride with the optimal epoch selection stride 1 (Sys.3 vs. Sys.4-5) outperformed the others selection methods (Sys.3 vs. Sys.6-7). More specifically, the best performing fine-tuned BERT model produced features with merging over the last three update epochs outperformed the baseline BERT features without fine-tuning and model merging by 11.36\% absolute on the CV data (Sys.3 vs. Sys.1). A more balanced performance across the CV and test sets were also obtained. Similar trends of accuracy improvements were also found on the fine-tuned Roberta features with model combination (Sys.9 vs. Sys.8). Based on these results, BERT and Roberta text encoder fine-tuning with model combination over the last update snapshots are used in all the following experiments of this paper.

\setcounter{magicrownumbers}{0}
\begin{table}[t]
	\setlength{\abovecaptionskip}{0.1cm}
	\centering
	\caption{AD detection performance evaluated by 10-fold CV on the ADReSS training set and test set using BERT/Roberta as the text feature extractor w/o model combination based fine-tuning of Fig. 1(A) using SVM classifier. All BERT/Roberta models fine-tuned on ground truth transcript (GT). All back-end classifiers trained and evaluated on ground truth transcript (GT). Sys.3-7 and Sys.9 show the effects of various update epochs selection schemes for model combination in Section 3. Random result (Sys.6) is averaged from 10 random selections. Fine-tuning without merging result (Sys.2) is averaged from performances of the selected single snapshot models.} 
	\label{tab:1}
	\centering
	\setlength{\tabcolsep}{1mm}{
	\scalebox{0.65}{
		\begin{tabular}{c|c|c|c|c|c|c|c}
			\noalign{\hrule height 0.5pt}
			\multirow{2}{*}{\textbf{Sys.}}
			&\multirow{2}*{{\begin{tabular}{c}
	            Feature
	            \end{tabular}}}
			&\multirow{2}*{{\begin{tabular}{c}
	            Fine-tuning \\ Data 
	            \end{tabular}}}
			&\multirow{2}*{{\begin{tabular}{c}
	            Comb \\-ination
	            \end{tabular}}}
            &\multirow{2}*{{\begin{tabular}{c}
	            Epoch \\ selection
	            \end{tabular}}}
	        &\multirow{2}*{{\begin{tabular}{c}
	            Epochs \\ merged
	            \end{tabular}}}
			&\multirow{2}*{{\begin{tabular}{c}
	            CV \\ Acc.(\%)
	            \end{tabular}}}
			&\multirow{2}*{{\begin{tabular}{c}
	            Test \\ Acc.(\%)
	            \end{tabular}}}
			\\
			&&&&&&  & 
			\\
			\noalign{\hrule height 1.0pt}
			\rownumber
			&\multirow{7}{*}{BERT}
			& N/A
			& $\times$
			& -
			& -
			& 74.75 & 87.50		\\
			\cline{3-8}
			\rownumber
			&&\multirow{6}{*}{GT}
			& $\times$
			& fixed stride(1) & [28,29,30] & 78.38 & 84.70\\
			\cline{4-8}
			\rownumber
			&&
			& \multirow{5}{*}{$\surd$}
			& fixed stride(1) & [28,29,30]
			& \bf{86.11}  & \bf{85.42} 			\\
			\cline{5-8}
			\rownumber
			&&&& fixed stride(3) & [24,27,30] & 86.11 & 79.17
			
			\\
			\cline{5-8}
			\rownumber
			&&&& fixed stride(10) & [10,20,30] & 84.26 & 81.25
			\\
			\cline{5-8}
			\rownumber
			&&&& random  & -  & 81.39 & 79.58
			\\
			\cline{5-8}
			\rownumber
			&&&& geo. interval & [18, 27, 30] & 82.41 & 83.33
			\\
			\cline{2-8}
			\rownumber
			&\multirow{2}{*}{Roberta}
			& N/A
			&  $\times$
			& -
			& -
			& 70.79	& 70.83 		\\
			\cline{3-8}
			\rownumber
			&& \multirow{1}{*}{GT}
			& \multirow{1}{*}{$\surd$}
			& fixed stride(1) & [28,29,30]
			&80.56 & 85.42		\\
			\noalign{\hrule height 1.0pt}
	\end{tabular}}}
	\vspace{-5mm}
\end{table}
\vspace{-0.2cm}
\section{Feature and Classifier Combination}
\vspace{-0.1cm}
This section investigates the feature combination of two text encoders (BERT and Roberta), and the back-end classifiers combination, with the manual transcriptions as inputs.

\setcounter{magicrownumbers}{0}
\begin{table*}[htbp]
	\vspace{-10mm}
	\setlength{\abovecaptionskip}{0.1cm}
	\centering
	\caption{ Accuracy of AD detection using different back-end classifiers with or without feature and/or model combination on manual transcripts, or one of three ASR systems (CNN-TDNN, Conformer, or their two-pass decoding combination \cite{tianzi2022conformer} “CNN-TDNN $\xrightarrow{}$ Conformer”) produced transcripts. “CNN-TDNN + Conformer” refers to separate AD detection using each ASR system’s specific transcripts before final decision voting combination. “LDA” stands for linear discriminant analysis, “GP” for Gaussian Process, “MLP” for multilayer perceptron, and “XGB” for extreme gradient boosting.}
	\label{tab:2}
	\centering
	\setlength{\tabcolsep}{1mm}{
	\scalebox{0.7}{
		\begin{tabular}{c|c|c|c|c|c|c|c|c|c}
			\noalign{\hrule height 0.5pt}
			\multirow{2}{*}{\textbf{Sys.}}
			&\multirow{2}{*}{{\begin{tabular}{c}
	            Feature
	            \end{tabular}}}
			&\multirow{2}{*}{{\begin{tabular}{c}
	            Combination \\ Method 
	            \end{tabular}}}
            &\multirow{2}{*}{{\begin{tabular}{c}
	            Classifier
	            \end{tabular}}}
	        &\multicolumn{2}{c}{{\begin{tabular}{c}
	            Manual \\ WER 0\%
	            \end{tabular}}} \vline
			&{\begin{tabular}{c}
	            CNN-TDNN \\ WER 30.69\%
	            \end{tabular}}
			&{\begin{tabular}{c}
	            Conformer \\ WER 26.70\%
	            \end{tabular}}
	       &{\begin{tabular}{c}
	            CNN-TDNN $\xrightarrow{}$ \\Conformer\\WER 26.50\%
	            \end{tabular}}
	        &{\begin{tabular}{c}
	            CNN-TDNN + \\ Conformer \\WER N/A
	            \end{tabular}}
			\\
			\cline{5-10}
            &&&& CV Acc.(\%) & Test Acc.(\%)& Test Acc.(\%)& Test Acc.(\%)& Test Acc.(\%) & Test Acc.(\%) \\
             
			\noalign{\hrule height 1.0pt}
			\rownumber
			&\multirow{7}{*}{BERT}
			& \multirow{5}{*}{N/A}
			& \multirow{1}{*}{SVM}
			& 86.11	& 85.42	& 85.42	& 81.25	& 81.25	& 81.25\\
			\cline{4-10}
			\rownumber
			&&
			& \multirow{1}{*}{LDA}
			& 86.11	& 79.17	& 72.92	& 72.92	& 81.25	& 72.92\\
			\cline{4-10}
			\rownumber
			&&
			& \multirow{1}{*}{GP}
			& 85.19	& 81.25	& 79.17	& 85.42	& 81.25	& 83.33
			\\
			\cline{4-10}
			\rownumber
			&&
			& \multirow{1}{*}{XGB}
			& 84.26	& 79.17	& 79.17	& 70.83	& 79.17	& 79.17
			\\
			\cline{4-10}
			\rownumber
			&&
			& \multirow{1}{*}{MLP}
			& 84.26 & 81.25 & 79.17 & 79.17 & 81.25 & 83.33
			\\
			\cline{3-10}
			\multirow{1}{*}{\rownumber}
			&&\multirow{1}{*}{{\begin{tabular}{c}
	            Classifier Best Selection
	            \end{tabular}}}
			& \multirow{1}{*}{{\begin{tabular}{c}
	            SVM + LDA + GP + XGB
	            \end{tabular}}}
			& \multirow{1}{*}{87.04 }
			& \multirow{1}{*}{85.42 }
			& \multirow{1}{*}{79.17 }
			& \multirow{1}{*}{81.25 }
			& \multirow{1}{*}{81.25}
			& \multirow{1}{*}{83.33 } 
			\\
			\cline{2-10}
			\rownumber
			&\multirow{7}{*}{Roberta}
			& \multirow{5}{*}{N/A}
			& \multirow{1}{*}{SVM}
			& 80.56 & 85.42 & 87.50 & 81.25 & 83.33 & 85.42\\
			\cline{4-10}
			\rownumber
			&&
			& \multirow{1}{*}{LDA}
			& 82.41 & 85.42 & 81.25 & 75.00 & 81.25 & 81.25
			\\
			\cline{4-10}
			\rownumber
			&&
			& \multirow{1}{*}{GP}
			&79.63 & 87.50 & 81.25 & 81.25 & 83.33 & 83.33
			\\
			\cline{4-10}
			\rownumber
			&&
			& \multirow{1}{*}{XGB}
			& 81.48 & 85.42 & 83.33 & 85.42 & 77.08 & 87.50
			\\
			\cline{4-10}
			\rownumber
			&&
			& \multirow{1}{*}{MLP}
			& 84.26 & 89.58 & 77.08 & 79.17 & 79.17 & 83.33
			\\
			\cline{3-10}
			\multirow{1}{*}{\rownumber}
			&&\multirow{1}{*}{{\begin{tabular}{c}
	            Classifier Best Selection
	            \end{tabular}}}
			& \multirow{1}{*}{{\begin{tabular}{c}
	            LDA + XGB  + MLP
	            \end{tabular}}}
			& \multirow{1}{*}{84.26 }
			& \multirow{1}{*}{\bf{91.67}}
			& \multirow{1}{*}{81.25 }
			& \multirow{1}{*}{79.17 }
			& \multirow{1}{*}{83.33 }
			& \multirow{1}{*}{81.25 }
			\\
			\cline{2-10}
			\rownumber
			&\multirow{7}{*}{{\begin{tabular}{c}
	            Bert + \\ Roberta
	            \end{tabular}}}
			& \multirow{5}{*}{{\begin{tabular}{c}
	            Feature \\ decision \\ voting
	            \end{tabular}}}
			& \multirow{1}{*}{SVM}
			& \bf{87.96} & 87.50 & \bf{93.75} & \bf{85.42} & \bf{87.50} & \bf{91.67}
			\\
			\cline{4-10}
			\rownumber
			&&
			& \multirow{1}{*}{LDA}
			&86.11 & 87.50 & 83.33 & 72.92 & 85.42 & 85.42
			\\
			\cline{4-10}
			\rownumber
			&&
			& \multirow{1}{*}{GP}
			& 85.19 & 83.33 & 83.33 & 83.33 & 85.42 & 87.50
			\\
			\cline{4-10}
			\rownumber
			&&
			& \multirow{1}{*}{XGB}
			& 86.11 & 81.25 & 85.42 & 83.33 & 87.50 & 87.50
			\\
			\cline{4-10}
			\rownumber
			&&
			& \multirow{1}{*}{MLP}
			& 87.04 & 85.42 & 81.25 & 85.42 & 81.25 & 81.25
			\\
			\cline{3-10}
			\multirow{1}{*}{\rownumber}
			&& \multirow{1}{*}{{\begin{tabular}{c}
	            Feature + Classifier.
	            \end{tabular}}}
	        & \multirow{1}{*}{{\begin{tabular}{c}
	            LDA + GP + MLP
	            \end{tabular}}} 
	        & \multirow{1}{*}{87.96}
	        & \multirow{1}{*}{85.42}
	        & \multirow{1}{*}{79.17 }
	        & \multirow{1}{*}{83.33 }
	        & \multirow{1}{*}{81.25 }
	        & \multirow{1}{*}{83.33}
			\\
            \cline{3-10}
			\multirow{1}{*}{\rownumber}
			&& \multirow{1}{*}{{\begin{tabular}{c}
	            Feature concat.
	            \end{tabular}}}
	        & \multirow{1}{*}{{\begin{tabular}{c}
	            SVM
	            \end{tabular}}} 
	        & \multirow{1}{*}{86.11}
	        & \multirow{1}{*}{81.25}
	        & \multirow{1}{*}{83.33}
	        & \multirow{1}{*}{79.17}
	        & \multirow{1}{*}{87.50}
	        & \multirow{1}{*}{81.25}
			\\
			\noalign{\hrule height 1.0pt}
	\end{tabular}}}
	\vspace{-4mm}
\end{table*}

\vspace{-0.3cm}
\subsection{Feature Combination}
\vspace{-0.1cm}
The two types of fine-tuned text embedding DNNs, BERT and Roberta, can capture linguistic information with discrepancy. To further reduce the uncertainty underlying the resulting text embedding features, we exploited a late feature fusion via majority voting over decisions made by a single type of classifier based on either BERT and Roberta features respectively (Feature decision voting, Table \ref{tab:2}), which is shown in Figure \ref{fig:AD_detection} (B, top right). The effectiveness of feature decision voting with different types of back-end classifiers is examined. We also explored alternative methods of fusing the two text embedding features by concatenation before feeding them into SVMs.

Two trends are observed from the manual transcripts based results in Table \ref{tab:2} (column 5-6). First, the method of feature fusion by decision-level majority voting produced improvements using all 5 back-end classifiers on the CV data over the comparable baselines using either BERT or Roberta features alone  (Sys.13-17 vs. Sys.1-5 \& Sys.7-11). On the test set, similar improvements were obtained for SVM and LDA classifiers (Sys.13 vs. Sys.1\&7, Sys.14 vs. Sys.2\&8). Second, the decision-level fusion by majority voting performed better than early fusion by concatenation (Sys.13 vs. Sys.19, col.5-6, Table \ref{tab:2}). Lastly, decision-level fusion among BERT and Roberta feature based SVM classifiers produced the best feature combination accuracy of 87.96\% on the CV data, and 87.50\% on the test (Sys.13 vs. 14-17, col.5-6, Table 2).

\vspace{-0.3cm}
\subsection{Classifiers Combination}
\vspace{-0.1cm}
Apart from the feature extractor stage, uncertainty could also manifest in architecture design and parameter estimation of back-end AD detection models on limited AD detection training data. To this end, we conducted classifier combination with majority voting over the binary AD detection decisions by a range of BERT (or Roberta) feature based back-end classifiers, as illustrated in Figure \ref{fig:AD_detection} (grey, bottom right). All possible feature-classifier combinations involving both the full set $2 \times 5$ combination and its various subsets were examined to maximize the diversity and complementarity among systems. The best performing feature-classifier combination configuration involves fine-tuned (with snapshot level model voting of Section 3) BERT or Roberta features trained LDA, GP and MLP classifiers (Sys. 18, col.5-6, Table \ref{tab:2}).  

The hyperparameters in the back-end classifiers are optimized by greedy search on CV, as mentioned in Section 2.2. More specifically, we used the SVM setting as in the baseline, LDA solved by SVD, GP with kernel $4.0 * RBF(5.0)$, XGB based on tree models with the number of boosting round set as 16, the maximum tree depth for base learners set as 2, learning rate $\eta = 0.4$, minimum loss reduction required to make a partition on the tree $\gamma = 0.1$ , and MLP with a single hidden layer of 256-dimension, L2 penalty parameter $\alpha = 1e-5$, optimized by limited memory BFGS. All other hyper-parameters were kept as default settings.

As shown in Table \ref{tab:2}, the first general trend is that classifier combination system consistently surpasses, or at least matches, the performance of any single classifier system across the CV and test sets (Sys.6 vs. Sys.1-5, Sys.12 vs. Sys.7-11, col.5-6, Table \ref{tab:2}). Secondly, conducting classifier combination on the top of feature combination produced no further improvements over pure feature combination (Sys.18 vs. Sys.13, col.5-6, Table \ref{tab:2}). This suggests further cross-system diversity and complementarity need to be injected into individual back-end classifiers’ design and training when feature combination is used.
\vspace{-0.5cm}
\section{ASR Output Based AD Detection}
\vspace{-0.1cm}
In this Section we apply the feature combination and classifier combination methods of Section 4 to AD detection using ASR system generated transcripts.
\vspace{-0.2cm}
\subsection{AD Detection Using Single ASR System's Output}
\vspace{-0.1cm}
The AD detection performance using various feature and classifier combination configurations for a single ASR system produced transcripts are shown in Table \ref{tab:2} (col.7-8). Several trends can be found. First, consistent with the previous trends on manual speech transcripts using feature combination (col.5-6, Table \ref{tab:2}), decision-level feature fusion by majority voting produced improvements using all 5 back-end classifiers on the test data over the comparable baselines using either BERT or Roberta features alone (Sys.13-17 vs. Sys.1-5 \& Sys.7-11, col.7-8, Table \ref{tab:2}). Second, the BERT+Roberta feature combination approach based SVM classifier (Sys.13), among all the combination strategies being considered in the same table, consistently produced the best AD detection accuracy using the two single ASR systems’ outputs (Sys.13 vs. Sys.1-12 \& 14-19, col.7-8, Table \ref{tab:2}). In particular, using the CNN-TDNN ASR system produced speech transcripts, this decision-level feature combination based SVM system produced the best published speech recognition outputs based AD detection accuracy of 93.75\% on the benchmark ADReSS test set. 
 
\vspace{-0.2cm}
\subsection{AD Detection Using Multiple ASR Systems' Outputs}
\vspace{-0.1cm}
 Compared with ASR performance measured on non-aged, healthy speech \cite{luscher2019rwth, tuske2021limit}, significantly higher speech recognition error rates are often found on elderly speech data \cite{pan2021using, gosztolya2019identifying, li2021comparative, ye2021development, geng2022speaker, vipperla2010ageing, rudzicz2014speech, zhou2016speech}. In order to mitigate the impact of possible ASR transcript errors on the downstream AD detection task, ASR system combination approaches are also investigated in this paper to account for the uncertainty over of the quality of speech recognition system outputs produced by the hybrid CNN-TDNN and end-to-end Conformer ASR systems considered in this paper. In this regard, we studied the efficacy of two approaches to combine the hybrid CNN-TDNN and end-to-end Conformer ASR systems. One is detection based on their two-pass decoding combined system produced outputs \cite{tianzi2022conformer} ("CNN-TDNN $\xrightarrow{}$ Conformer", Table \ref{tab:2}), while the other performs separate AD detection using either of the two ASR system’s output transcripts before final decision voting combination ("CNN-TDNN+Conformer", Table \ref{tab:2}). 

From results presented in Table 2 (col.9-10), we first confirmed that BERT plus Roberta feature decision voting system with the SVM classifier is again the best performing system among all (Sys.13 vs. Sys.1-12 \& 14-19, col.9-10, Table \ref{tab:2}), regardless of which ASR combination approach being used. However, using neither of the two ASR system combined outputs led to improvements over using the hybrid CNN-TDNN system transcripts only (Sys. 13, col.9-10 vs. col.7). This suggests further diversity and complementarity may be required in ASR system combination and integration with the downstream AD detection component.
The performance of the best performing AD detection system in this paper (sys. 13, Table 2) are further contrasted in Table 3 with those that are obtained on the same ADReSS20 challenge task and report in the recent literature to demonstrate their competitiveness.
\vspace{-0.35cm}

\setcounter{magicrownumbers}{0}
\begin{table}[t]
	\setlength{\abovecaptionskip}{0.1cm}
	\centering
	\caption{ Comparison of test results (rounded to one decimal place) against the SOTA based on different modality for the ADReSS20 challenge. "Pre.", "Rec." and "F1" represents precision, recall and F1 scores regarding AD as the relevant class.}
	\label{tab:3}
	\centering
	\setlength{\tabcolsep}{1mm}{
	\scalebox{0.7}{
		\centering
\begin{tabular}{c|c|cccc}
\hline
\multirow{2}{*}{Study} & \multirow{2}{*}{Modality} & \multicolumn{4}{c}{Test}                                                                               \\ \cline{3-6} 
                       &                           & \multicolumn{1}{c|}{Acc.(\%)} & \multicolumn{1}{c|}{Pre.(\%)} & \multicolumn{1}{c|}{Rec.(\%)} & F1(\%) \\ \noalign{\hrule height 1.0pt}
Syed et al. \cite{syed2021automated} & Text (Manual)            & \multicolumn{1}{c|}{91.7}     & \multicolumn{1}{c|}{-}        & \multicolumn{1}{c|}{91.7}     & -      \\ \hline
Li et al. \cite{li2021comparative} & Text (ASR)               & \multicolumn{1}{c|}{88.0}     & \multicolumn{1}{c|}{91.0}     & \multicolumn{1}{c|}{83.0}     & 87.0   \\ \hline
Martinc et al. \cite{martinc2021temporal} & Audio + Text (Manual)    & \multicolumn{1}{c|}{93.8}     & \multicolumn{3}{c}{-}                                                  \\ \hline
Laguarta et al. \cite{laguarta2021longitudinal} &
  \begin{tabular}[c]{@{}c@{}}Audio +\\ pre-trained biomarkers\tablefootnote{They transferred knowledge from pre-trained detection system using biomarkers (e.g. cough features) developed with non-public data} \end{tabular} &
  \multicolumn{1}{c|}{93.8} &
  \multicolumn{3}{c}{-} \\ \hline 
\multirow{2}{*}{{\begin{tabular}{c}Our work \\ (Sys.13, Tab. 2)\end{tabular}}} & Text (Manual) & \multicolumn{1}{c|}{91.7} & \multicolumn{1}{c|}{91.7} & \multicolumn{1}{c|}{91.7} &  91.7 \\ 
 & Text (ASR) & \multicolumn{1}{c|}{93.8} & \multicolumn{1}{c|}{92.0} & \multicolumn{1}{c|}{95.8} & 93.9 \\ \noalign{\hrule height 1.0pt}
\end{tabular}
    }}
	\vspace{-5mm}
\end{table}

\section{Conclusion}
\vspace{-0.1cm}
The development of a state-of-the-art speech-based AD detection system constructed using the ADReSS20 challenge dataset was presented in this paper. AD detection was performed based on either manual or ASR system transcripts. A series of model ensemble approaches were used in fine-tuning of BERT and Roberta pre-trained text encoders, text  embedding feature and back-end classifier combination, as well as the integration with front-end ASR system to address the modeling uncertainty issue resulted from the sparsity of AD detection training data. Experiments conducted on the ADReSS20 Challenge dataset suggest consistent performance improvements were obtained using model and feature combination in AD detection system development. 
Tighter integration between the ASR system and AD detection components will be investigated in future research. 

\vspace{-0.35cm}
\section{Acknowledgements}
\vspace{-0.2cm}
This research is supported by Hong Kong RGC GRF grant No.
14200021, 14200218, 14200220, TRS T45-407/19N and Innovation \& Technology Fund grant No. ITS/254/19.

\bibliographystyle{IEEEtran}
\vspace{-0.2cm}
\bibliography{mybib}
\vspace{-0.1cm}

\end{document}